\documentclass{article}
\usepackage{acra}
\usepackage{graphicx}
\usepackage{booktabs}
\usepackage{amsmath}
\usepackage{hyperref}

\title{Deep Reinforcement Learning for Local Path Following of an Autonomous Formula SAE Vehicle}

\begin{document}

\author{Harvey Merton$^1$, Thomas Delamore, Karl Stol \and  Henry Williams$^2$\\
        Centre for Automation and Robotic Engineering Science \\ 
        The University of Auckland, NZ \\
        $^{1}$hmer101@mit.edu, $^{2}$henry.williams@auckland.ac.nz}

\maketitle

\begin{abstract}
    With the continued introduction of driverless events to Formula:Society of Automotive Engineers (F:SAE) competitions around the world, teams are investigating all aspects of the autonomous vehicle stack. This paper presents the use of Deep Reinforcement Learning (DRL) and Inverse Reinforcement Learning (IRL) to map locally-observed cone positions to a desired steering angle for race track following. Two state-of-the-art algorithms not previously tested in this context: soft actor critic (SAC) and adversarial inverse reinforcement learning (AIRL), are used to train models in a representative simulation. Three novel reward functions for use by RL algorithms in an autonomous racing context are also discussed. Tests performed in simulation and the real world suggest that both algorithms can successfully train models for local path following. Suggestions for future work are presented to allow these models to scale to a full F:SAE vehicle.
\end{abstract}

\section{Introduction}
\begin{figure}[h]
        \centering
        \includegraphics[width = 0.7\linewidth]{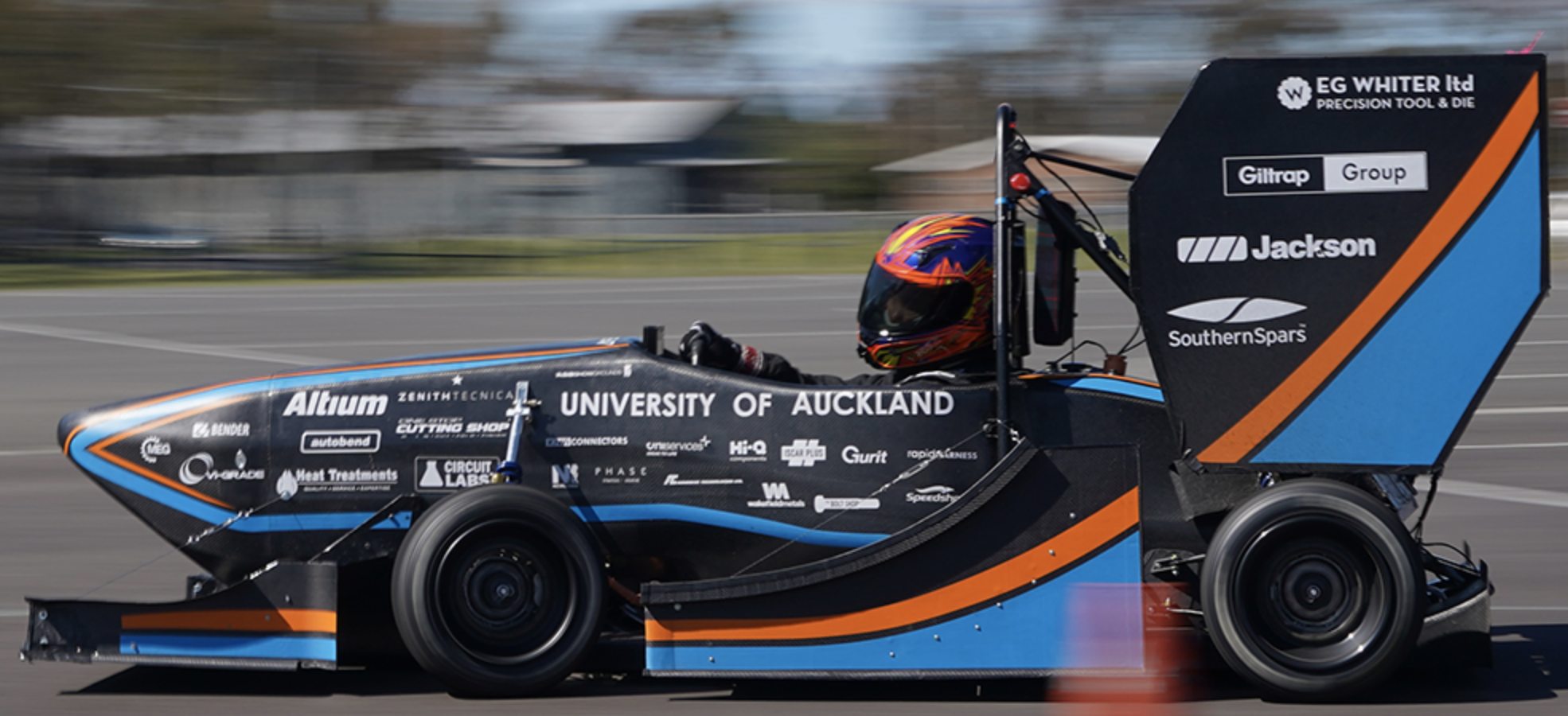}
        \caption{Auckland University F:SAE Team's 2019 electric vehicle.}
        \label{fig:fsae47_2019}
    \end{figure}

    Formula:Society of Automotive Engineers (F:SAE), also known as Formula Student (FS) in some regions, is a series of international events that challenge university teams to design, manufacture and race a fully custom, formula-style race car (see Figure \ref{fig:fsae47_2019}).
    
    In 2017, Formula Student Germany (FSG) became the first FS competition to introduce a driverless vehicle class - called Formula Student Driverless (FSD) (see \cite{FSGRulesAll} for 2023 rules). Vehicles in this class must autonomously navigate tracks delineated with blue cones on the left side and yellow on the right.  
    
    The current dominant approach to autonomous driving is highly modular: cone perception by camera and/or lidar, a variant of SLAM for localisation and mapping, followed by motion planning and control, often with model predictive control (MPC) \cite{Kabzan2019}. Although these methods have found success in some constrained and deterministic environments, they often tend to struggle when it comes to uncertain, dynamic environments \cite{Gonzalez2016}. 

    Reinforcement learning (RL) is one of the three fundamental machine learning (ML) approaches, alongside supervised and unsupervised learning, and has been of research interest since the late 20th century \cite{Kaelbling1996}. RL varies from the other approaches in that the algorithms autonomously collect data from an environment to train models, rather than being provided with training data. They determine policies during training by maximising a reward obtained from their environment, which is determined by a reward function. When the policy trained by RL is stored in a deep neural network (rather than a table or other method), it is known as deep reinforcement learning (DRL) \cite{Mnih2013}.
    
    Since 2016, deep reinforcement learning (DRL) has rapidly gained popularity for robotic navigation problems. As discussed in \cite{Zhu2021}, DRL has found more success than traditional methods in local obstacle avoidance, indoor navigation, multi-robot navigation and social navigation. It is also noted that DRL is able to achieve various other navigation tasks without an intermediate map representation and in many cases, can relate image inputs directly to control outputs in `end-to-end' ML \cite{levine2016end}. These results showcase DRL's ability to deal with the uncertain and dynamic environments that model-based methods struggle in. It is therefore no surprise that DRL has begun to find success in solving various problems in the full autonomous driving stack \cite{Jaritz2018} \cite{Wayve2021}.
    
    This paper focuses on the feasibility of using reinforcement learning to allow local path following of an FSD race track. The main contributions are a comparison of two leading RL and inverse RL algorithms applied to racing in simulation, an evaluation of these algorithms' ability to transfer to the real world, and a discussion of the training setup that results in the shortest convergence time. Three novel reward functions for use by RL algorithms in an autonomous racing context are also presented.

\section{Background and Related Work}
    A huge variety of RL algorithms exist, each of which have their own advantages across different tasks and domains \cite{Arulkumaran2017} \cite{Wang2022}. This paper specifically focuses on the comparison of forward and inverse reinforcement learning (IRL), so discussion here is limited to that distinction. 

    Unlike RL algorithms' self-collection of training data, Imitation Learning (IL) algorithms attempt to imitate an expert's actions \cite{Piot2017}. Two important sub-classes of IL are Behaviour Cloning (BC) and IRL. BC algorithms attempt to directly mimic an expert's behaviour, whereas IRL methods attempt to discover a reward function from demonstrated `expert' actions \cite{Ng2000}. IL algorithms are useful as they don't require user-defined reward functions which are difficult to define. The disadvantage is that they do require an expert to demonstrate actions, which is not always available.
    
    \textbf{Soft Actor Critic (SAC)} is a continuous action space, model-free, off-policy, actor-critic RL algorithm \cite{Haarnoja2018}. Because SAC is off-policy, a behaviour generator is used to generate somewhat stochastic actions in an environment. The reward and observation returned by an environment are passed to the critic network in the agent. Outputs from the critic are then used to update ANN weights in the actor and critic using an optimiser. 
    
    As discussed in  \cite{Haarnoja2018}, SAC is more robust to varied ANN hyperparameters and initialisation seeds across a range of simulation environments than the deep deterministic policy gradient (DDPG), proximal policy optimisation (PPO), soft Q-learning (SQL) and twin delayed deep deterministic policy gradient (TD3) algorithms. This is due to the ``maximum entropy" approach where the behaviour generator generates many random actions to better explore the environment \cite{Haarnoja2018}. For these reasons, SAC is selected as the RL algorithm for investigation in this paper.
    
    \textbf{Adversarial Inverse Reinforcement Learning (AIRL)} is a model-free, deep inverse reinforcement learning algorithm (IRL) first presented in \cite{Fu2017}. Apart from the expert buffer, the biggest change in AIRL vs SAC is that no reward is required from the environment. The reward is instead replaced by a ``discriminator" ANN that attempts to tell the difference between expert observation and the real observation generated by the same action. 
    
    AIRL shows mean rewards in a range of simulated environments that are at least comparable to other IL methods, such as generative adversarial imitation learning (GAIL) and adversarial network guided cost learning (GAN-GCL) \cite{Fu2017}. Further, AIRL was shown to adapt to variation in the underlying domain more readily than other similar algorithms, which is ideal when trying to transfer from simulation to reality such as in this paper. For these reasons, AIRL is selected as the IRL algorithm for investigation in this paper.

    \subsection{Deep Reinforcement Learning and Autonomous Vehicles}       
        The survey conducted in \cite{Kiran2021} focuses specifically on deep RL (DRL) methods for autonomous driving. It notes that various DRL algorithms have been successfully applied to vehicle control \cite{Xia2016}, dynamic path planning \cite{Aradi2022} and trajectory optimization \cite{Isele2018} problems. Recent work also discusses that DRL methods have the potential to be particularly effective in dynamic and unpredictable environments. One such environment is autonomous overtaking, where traditional methods usually fail \cite{Song2021}. 
        
        Several challenges related to DRL methods are identified in \cite{Kiran2021} including: validating performance, covering the simulation-reality gap, sample efficiency, designing good reward functions and ensuring safety. It is stressed that, like in \cite{Zhu2021}, most DRL methods presented are only proven in simulation and haven't been tested in the more complex real world.
    
    \subsection{Autonomous Vehicles in F:SAE}
        The University of Auckland has published some initial results on using DRL for path following \cite{salvaji2023racing}. They compare Deep Q Networks (DQN) for discrete action spaces with Twin Delayed DDPG (TD3) for continuous action spaces on a differential drive turtlebot. Although this is a good start, the research is limited by the turtlebot platform not being representative of an Ackermann-steering-based F:SAE vehicle. This is a problem that must be addressed both in simulation and the real world. Further, the research did not consider IL methods, and suggested that more complex reward functions be attempted. This paper addresses the previous problems whilst also testing a different DRL algorithm.
        
        To our knowledge, the only other paper that discusses the use of a RL-related technique in an F:SAE context is \cite{Zadok2019}. This paper describes the use of an IL algorithm with a modified PilotNet artificial neural network (ANN) for end-to-end autonomous driving. To train the model, data was gathered by human drivers in model F:SAE vehicles using the photo-realistic AirSim simulation software. However, this method is limited because the autonomous vehicle can never outperform the human drivers. Further, human drivers are required for data collection and the parameters of the vision module cannot be altered independently of the DRL model.    

\section{Simulation Setup and Results}
    The objective of this research is to train a model using DRL (\emph{training} stage) that can allow a vehicle to reliably navigate a track in both simulations (see Figure \ref{fig:track_FSG}) and the real world (\emph{inference} stage). The full implementation using OpenAI gym can be found \href{https://github.com/HarveyMerton/uoa\_fsd}{here}. 
    
    \subsection{Simulation Setup} 
        Two \textbf{co-ordinate systems (CS)} are defined in Figure \ref{fig:fssim3D}: 
        
        \begin{itemize}
            \item  \{X, Y, Z\} - origin is attached to the base of the chassis in the centre of the driver's seat and X is aligned with the chassis.
            \item \{r, $\theta$\} (used for cone position noise) - same origin as \{X, Y, Z\} and $\theta = 0$ is aligned with the X axis.
        \end{itemize}
        
       A third, static, global CS \{x, y, z\} is defined as the \{X, Y, Z\} CS before the vehicle moves after an environment reset. The \{X, Y, Z\} CS is pictured in the pose defining the \{x, y, z\} CS in Figure \ref{fig:fssim3D}.
        
        The control inputs to the vehicle are in the form of a steering angle and throttle position, each normalised from -1 to 1. The steering angle input is limited to $\pm{0.4}$ (corresponding to $\pm 18$\,deg) as this reflects a physically realistic Ackermann steering range. During \emph{inference}, the steering angle is selected by the trained models based on cone pose inputs (relative to the \{X, Y, Z\} co-ordinate system seen in Figure \ref{fig:fssim3D}) from a [360 deg] 'vision sensor'. The vision sensor is modelled as a 10\,m radius around the vehicle within which cone positions and colours can be detected. The three cones closest to the camera on the left, and three closest on the right are used as inputs (the \emph{observation space}/the state of the vehicle in the environment) to the DRL algorithms. 
        
        The throttle position of the vehicle is controlled by a P controller to maintain a constant speed at 4\,m/s (14.4 km/h); a reasonable speed for a driverless vehicle's first lap of an unknown track. Thus, the \emph{action space} of the problem is a steering angle in normalised range $\pm{0.4}$.
        
        Uncertainty in cone position measurement is modelled with random Gaussian noise by the default values of $\mu_{r,\theta} = 0.0$, $\sigma_r = 0.2$ and $\sigma_{\theta} = 0.007$ ($\mu$ is mean and $\sigma$ is standard deviation in $r,\theta$ co-ordinates defined in Figure \ref{fig:fssim3D}). These values are representative of the noise observed by the cone detection system in \cite{Kabzan2019}. 
    
    \subsubsection{Simulation Environment} \label{sect:fssim}
    
        \begin{figure}[h]
            \centering
            \includegraphics[width=0.3\textwidth]{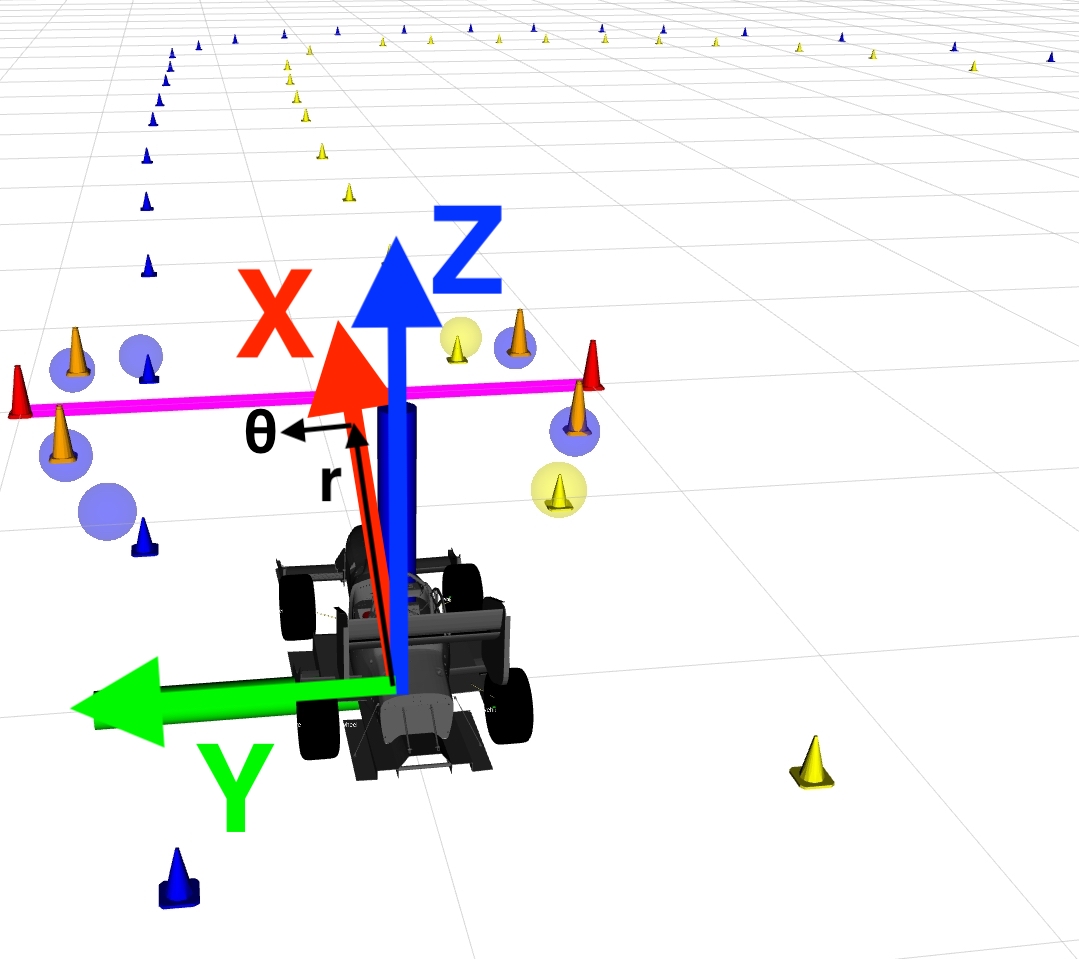}%
            \caption{FS simulator visualisation - RVIZ.}\label{fig:fssim3D}
        \end{figure} 
        
        The FS simulator (Figure \ref{fig:fssim3D}) was developed by the AMZ Driverless team to model a realistic Formula Student vehicle in the Robot Operating System (ROS). It is based on a vehicle model discretized through a forward Euler approximation \cite{Kabzan2019} (timestep size of 0.1\,s). The simulator uses a point mass with centre of mass position and front/rear weight balance, yaw inertia, tyre model (Pacejka's magic formula \cite{pacejka_bakker_1992}), basic aerodynamics model (whole vehicle lift and drag coefficients) and basic torque vectoring. No suspension modelling is included.
        
        \begin{figure}[htbp] 
            \centering
            \includegraphics[width=0.45\textwidth]{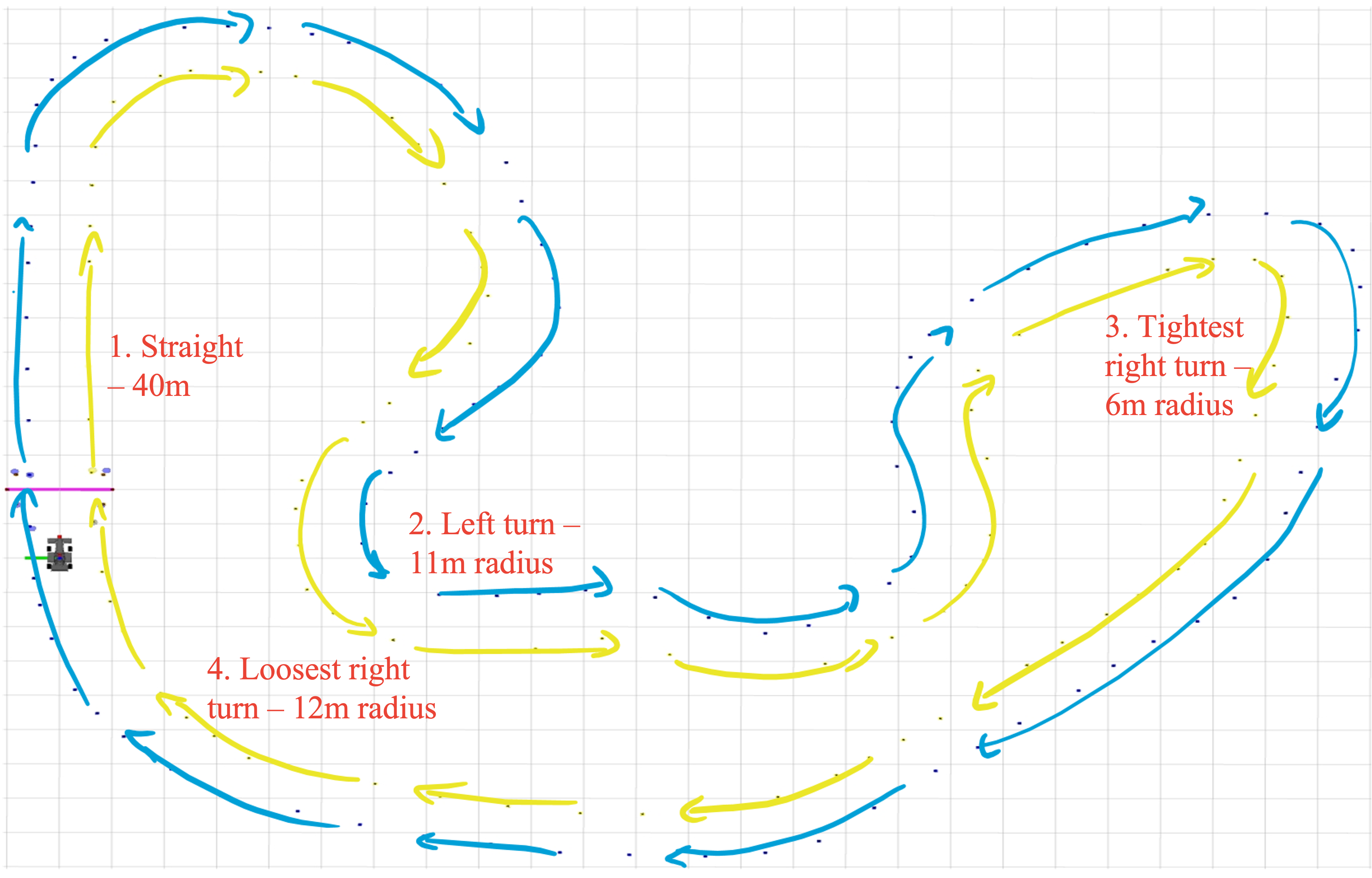}%
            \caption{FSG 2019 full-scale trackdrive track in simulation. \iffalse Note blue cones are represented with blue arrows, while yellow cones are represented with purple arrows for clarity. \fi}\label{fig:track_FSG}
        \end{figure}
        
        Figure \ref{fig:track_FSG} shows the track defined for training and inference testing in simulation. This track is representative of the 2019 FSG trackdrive track and so is named `FSG track'. When the vehicle travel direction and corresponding coloured cone sides are swapped, the track is called `inverse FSG', and represents an entirely new track as far as the vehicle is concerned. The track is approximately 400\,m in total length.
        
        One episode is the time from when the vehicle starts in the position shown in Figure \ref{fig:track_FSG} until all four vehicle wheels leave the cone-marked track (failure defined by FSG rules \cite{FSGRulesAll}) or the vehicle crosses the pink finish line for the second time (full lap completed). 
    
    \subsubsection{RL Algorithm Configuration}\label{sect:algos} 
        The SAC algorithm uses two fully-connected ANNs: one for the actor and one for the critic. These both have two hidden layers comprised of 256 nodes each, using the Tanh activation function. Both networks use a linear output activation function. SAC training uses a batch size of 256 and a replay buffer of size $10^6$. All networks in both SAC and AIRL use the Adam optimizer with a learning rate of $3e^{-4}$. The hyperparameters and network structures of both SAC and AIRL were determined through a series of brief initial tests. They were selected using the same settings seen in tests S1b and A4 in Table \ref{table:res_training}, and were the combinations that saw the fastest initial successful convergences. Given that the setup decisions discussed later have lead to functioning models, further work could be performed here to optimize these parameters. This may perhaps lead to faster and more stable convergence.
        
        The AIRL algorithm uses four ANNs: one for the actor, one for the critic and two for the discriminator. The actor and critic networks use the same network structure as SAC, except they only use 64 hidden nodes at each hidden layer. The discriminator networks also both have two hidden layers of 64 nodes each, but instead use the ReLU activation function. AIRL training uses a batch size of 64 and a serialized buffer to take in expert observations. The expert in simulation is simply a PID controller that moves the vehicle towards a hard-coded centre line of the training track (called \emph{pure pursuit}). Note that pure pursuit is only used in simulation as it is easy to generate a centerline around a track known a priori. An AIRL-trained model based on this is intended to have similar center-following behaviour but hopefully with increased robustness. The idea is that with a varying number of cones being detected (e.g. perhaps two cones on the same side of the track), finding a centerline to follow in the real world may be challenging. Using the AIRL-trained model means that edge-cases should be seen and accounted for during training rather than explicitly hard-coding responses in these cases.
        
        Three potential reward functions to be used by SAC are defined in Equations \ref{eq:rewardStep}-\ref{eq:rewardTime}. Each is determined through domain knowledge and different logical ways of describing a race car's progress around a track. These functions have some common symbols: $i$ represents time-step $i$, $\alpha_1$ through $\alpha_7$ are tunable scaling parameters, `!' is a logical `NOT', and $done_i$ is a binary variable which equals 1 if the episode has ended and 0 if not. $MAX$ is an upper limit placed on the reward able to be received at any one time-step which prevents reward explosion when dividing by small numbers. Note that exponential or logarithmic functions could be used in an attempt to discourage movement towards track boundaries or otherwise. These are avoided here to keep the functions as simple and understandable as possible, but could be explored in future work.

    \subsubsection{Reward function 1}
   
        \begin{equation}
        \begin{split}
            \label{eq:rewardStep}
            Reward_i & = \min \biggl( !done_i \times \biggl(\alpha_1 + \phantom{\frac{\alpha_2}{|ang_i - ang_{i-1}|}} \\
                     & \quad + \frac{\alpha_2}{|ang_i - ang_{i-1}|}\biggr), \,MAX \biggr)
        \end{split}
        \end{equation}
        
        

        where $ang_i$ is the steering angle at time-step $i$. Setting $\alpha_2 = 0$ in reward Function \ref{eq:rewardStep} yields the simplest possible reward function, intended to be easy for a learning algorithm to discover. A reward of $\alpha_1$ is given for every time-step that the vehicle remains between the cones, otherwise a reward of 0 is given. As a constant throttle value is sent to the simulator at all time-steps, the number of time-steps the vehicle achieves in an episode is roughly proportional to the distance it has travelled around the track. Thus, the cumulative reward received for a full episode is roughly proportional to the distance travelled around the track.

        Setting $\alpha_2$ to a non-zero value in reward Function \ref{eq:rewardStep} results in a higher reward given a smaller steering angle change (relative to the last time-step). This modification is intended to increase the `smoothness' of travel/decrease zigzagging around the track.
    
\begin{figure}[htbp] 
            \centering
            \includegraphics[width=0.5\textwidth]{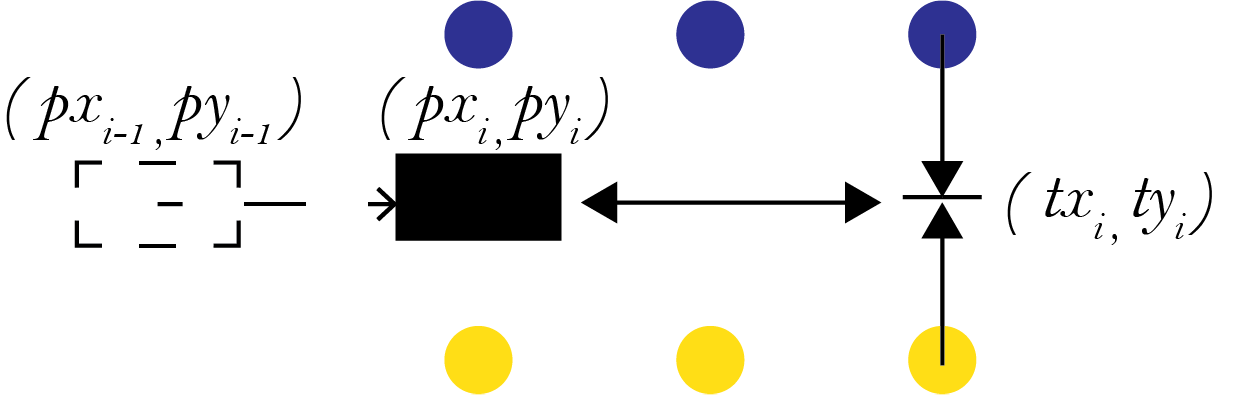}%
            \caption{Reward function 2 visualisation.}\label{fig:sim_fxn2}
        \end{figure}

    \subsubsection{Reward Function 2}
    
        \begin{equation}
        \begin{split}
            \label{eq:rewardDist}
            & Reward_i = done_i \times \alpha_3 + \\ 
                     & \quad + \min\left(!done_i \times \frac{\alpha_4}{\sqrt{(tx_i - px_i)^2 + (ty_i - py_i)^2}},  MAX\right)
        \end{split}
        \end{equation}
        
        where $tx_i$, $ty_i$, $px_i$ and $py_i$ are the target (x, y) and current (x, y) positions in the global CS \{x, y, z\} at time-step $i$ respectively. The target position is the midpoint between the furthest pair of blue and yellow cones seen in time-step $i-1$ (see Figure \ref{fig:sim_fxn2}); reward is given by how close the vehicle is to this point in the next time step. The intent of this function is to reward centerline following behaviour and prevent zig-zagging, but comes at the cost of increased complexity over Function \ref{eq:rewardStep}. Note that $\alpha_3$ can be used to give a negative reward when the vehicle exits the track.
        
        \subsubsection{Reward Function 3}
        \begin{equation}
        \begin{split}
            \label{eq:rewardTime}
            & Reward_i = done_i \times \alpha_5 + \\
            & + \min\left(!done_i \times \left(!target_i \times \alpha_6 + \frac{target_i \times \alpha_7}{c_i - c_t}\right), \,MAX\right)
        \end{split}
        \end{equation}
        
        where $c_i$ and $c_t$ are the total time-step counts (since episode start) at time-step $i$ and $t$ (time-step when the target was set) respectively. $target_i$ is a binary variable which equals 1 if the vehicle has reached a predefined target and 0 if not. The predefined target is the midpoint between the furthest pair of blue and yellow cones observed at time-step $c_t$ (similar to Figure \ref{fig:sim_fxn2}). Once the target is reached in time-step $c_i$, a new target is generated and $c_t$ is set to $c_i$.
        
        The target position in this function is similar to Function \ref{eq:rewardDist} except the same target is kept until it is reached, rather than moving the target each time a new set of cones can be observed. This allows the reward to be associated with time, thus linking the reward to the ultimate objective of minimising lap time. 
    
    \subsection{Training Results and Discussion}\label{sect:sim_train_results}
        Videos of the training and inference results in simulation and the real world can be found \href{https://www.youtube.com/watch?v=QyHlDjPPw_o}{here}.
        
        Table \ref{table:res_training} summarises how some of the key parameters affect the number of training episodes to convergence (when 5 consecutive full laps have been completed by the trained model). Key reported parameters include: `angle limit', `rate' and `colour ID'. These parameters represent if a  $\pm{18}^{\circ}$ steering angle limit, steering angle rate limit or cone colour identifiers are applied. `Rollout' length refers to the number of steps ahead the AIRL algorithm is looking when optimising the selected steering angle. 
        
        \begin{table}[ht]
        \centering
            \scalebox{0.7}{
            \begin{tabular}{l l r r r r}\toprule
                ID & What testing & Episodes to & Angle & Rate & Colour\\
                & \underline{\textbf{SAC}} & convergence & limit & ($^{\circ}/s$) & ID\\\midrule
                & \textbf{Function 1} & &\\
                S1b & $\alpha_1 = 1,\,\alpha_2 = 0$ & 504 & N & N & N\\
                S14 & $\alpha_1 = 1,\,\alpha_2 = 0$ & 362 & Y & N & N\\
                S2b & $\alpha_1 = 100,\,\alpha_2 = 0$ & 176 & N & N & N\\
                S3b & $\alpha_1 = 1,\,\alpha_2 = 1$ & 528 & Y & N & N\\\midrule
                
                & \textbf{Steering angle limits} & \multicolumn{4}{l}{($\alpha_1 = 1,\,\alpha_2 = 0$)} \\
                S1b & $\pm{45}^{\circ}$ & 504 & N & N & N\\
                S6 & $\pm{18}^{\circ}$ & 352 & Y & N & N\\\midrule 
                
                & \textbf{Colour identifiers} & \multicolumn{4}{l}{($\alpha_1 = 1,\,\alpha_2 = 0$)} \\
                S6 & Without & 352 & Y & N & N\\
                S15 & With & 351 & Y & N & Y\\\midrule 
                
                & \textbf{Rate limit (deg/s)} & \multicolumn{4}{l}{($\alpha_1 = 1,\,\alpha_2 = 0$)} \\
                S15 & None & 351 & Y & N & Y\\
                S12b & 90 & 1942* & Y & 90 & Y\\
                S18 & 112.5 & 735 & Y & 112.5 & Y\\
                S17 & 135 & 615 & Y & 135 & Y\\
                S16 & 180 & 624 & Y & 180 & Y\\\midrule 
                
                & \textbf{Cone pose uncertainty} & \multicolumn{4}{l}{($\alpha_1 = 1,\,\alpha_2 = 0$)}  \\
                S18 & Default & 735 & Y & 112.5 & Y\\
                S21 & None & 443 & Y & 112.5 & Y\\\midrule

                & \underline{\textbf{AIRL}} & & Rollout & Rate & Colour\\\midrule 
                & \textbf{Rollout length (steps)} & & &($^{\circ}/s$) & ID\\
                A2 & 50\,K & 3354* & 50\,K & N & Y\\
                A4 & 25\,K & 1356 & 25\,K & N & Y\\
                A5 & 12.5\,K & 752 & 12.5\,K & N & Y\\
                A3b & 6125 & 1277* & 6125 & N & Y\\
                A10 & 1000 & 121 & 1000 & N & Y\\
                A12 & 500 & 316 & 500 & N & Y\\
                A9 & 100 & 69** & 100 & N & Y\\\midrule
                
                & \textbf{Cone pose uncertainty} & & \\
                A10 & Default & 121 & 1000 & N & Y\\
                A11 & None & 1700 & 1000 & N & Y\\\midrule
                
                & \textbf{Rate limit (deg/s)} & & \\
                A10 & None & 121 & 1000 & N & Y\\
                A13 & 112.5 & 1386 & 1000 & 112.5 & Y\\\bottomrule
                
                & \multicolumn{5}{l}{* = Max episode (not converged), ** = unstable, Y = yes, N = no}\\\bottomrule
            
            \end{tabular}}
            \caption{Training results summary.}
            \label{table:res_training}
        \end{table}

        \underline{\emph{SAC}}\label{sect:training_results_sac}
        \textbf{Reward function:} 
        Comparing results S14 and S3b in Table \ref{table:res_training}, it can be seen that adding a non-zero $\alpha_2$ into reward function \ref{eq:rewardStep} appears to increase the overall convergence time from 362 to 528 episodes. Further, qualitative analysis of vehicle trajectory shows no improvement in ``smoothness" as desired. This is due to the backend ANN having no ``memory" of the previous steering angle, thus $\alpha_2 = 0$ going forward. 
        
        Comparing S1b with S2b it is seen that increasing the magnitude of $\alpha_1$ from 1 to 100 drastically reduces the number of episodes required before convergence. As this was discovered after most other training tests, $\alpha_1 = 1$ for most of the remaining tests.

        \textbf{Cone colour identifiers:} 
        Adding cone colour identifiers increases convergence time from 352 to 488 episodes. However, qualitative inference performance suggests a much straighter trajectory when identifiers are included. As the impact on training performance isn't overwhelming, colour identifiers are included going forward. 
        
        \textbf{Rate limit:}
        This prevents the RL algorithms causing physically impossible instantaneous angle changes. $112.5^{\circ}/s$ is the smallest rate limit applied that still allowed SAC to converge (compare S18 with S12b); this limit is applied going forward as it reflects a physically possible rate of steering angle change without extremely long convergence times. 
        
        \textbf{Cone pose uncertainty:}
        Comparing S18 with S21, removing the uncertainty in cone positions during training significantly reduces training time (735 to 443 episodes). However, to follow the literature concerning domain randomisation, cone uncertainty in training is retained to improve trained model robustness. Thus, \emph{S18 is selected as the SAC model to take forwards to inference testing}.
        
        \underline{\emph{{AIRL}}}
        \textbf{Rollout length:}
        Although A9 shows extremely fast convergence with 69 episodes, the performance quickly dropped off and no full laps were completed until episode 826, when convergence was again reached. This suggested unstable convergence and so a total amount of 1,000 steps was selected as a fast converging and stable rollout length to proceed with.
        
        \textbf{Cone pose uncertainty:}
        In the opposite manor to SAC, removing cone pose uncertainty actually increased convergence time from 121 episodes in A10 to 1,700 in A11. This is likely due to the expert data being collected with the cone pose uncertainty present. 
        
        \textbf{Rate limit:}
        For comparability with SAC and to allow physical implementation, the $112.5^{\circ}/s$ rate limit was carried forward despite the 11.5 times increase in convergence time. As such, \emph{A13 is selected as the AIRL model to take forwards to inference testing}.

    \subsection{Inference Results and Discussion}\label{sect:sim_inf_results}

        \begin{figure}[htbp] 
            \centering
            \includegraphics[width=0.5\textwidth]{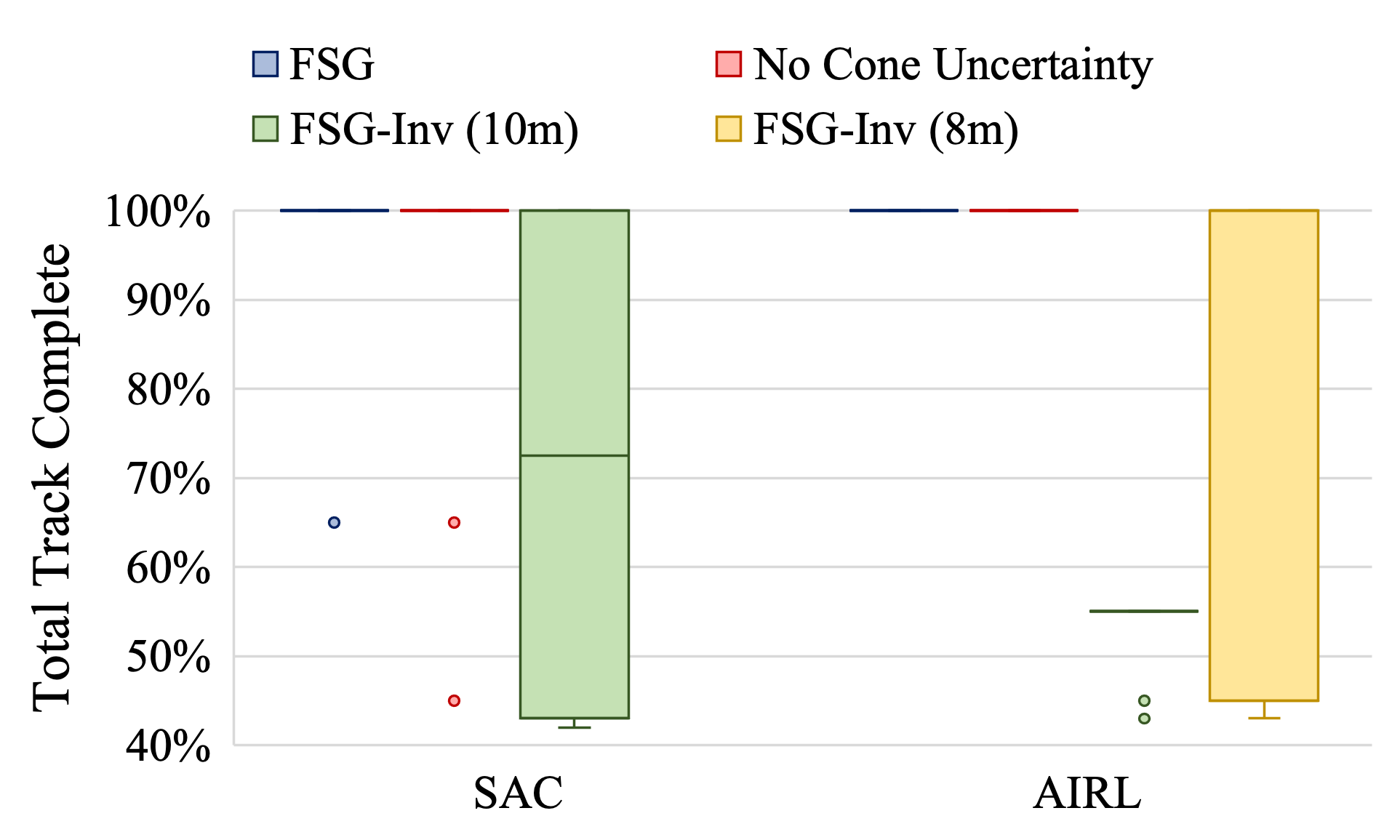}%
            \caption{SAC vs AIRL simulation inference performance - 120 trials per variant.}\label{fig:res_inf_sac_vs_airl}
        \end{figure}

    \subsubsection{FSG Track}
        As seen in Figure \ref{fig:res_inf_sac_vs_airl}, both SAC and AIRL-trained models achieve a median 100\% track completion on the FSG track. Similar performance is obtained when cone pose uncertainty is removed on the inference tracks suggesting that domain randomisation may have helped increase robustness to changing cone pose uncertainty. 

    \subsubsection{Inverse FSG Track}
        Figure \ref{fig:res_inf_sac_vs_airl} shows that the performance of both SAC and AIRL-trained models is worse on the inverse FSG track than the FSG track. This is expected as the inverse FSG track was not seen during training. The 72\% median track completion shown by the SAC model is misleading as the results are clustered in two groups: around 43\%\,-\,45\% and 100\%. The 43\%\,-\,45\% failure points corresponds to feature 3 in Figure \ref{fig:track_FSG}), creating a tighter left turn in the inverse track than present in the FSG track on which the model was trained. This shows that SAC is not able to generalise well to a corner this far out of its original training data. However, the 100\% completion clustering of 5 trials shows that the rest of the track can be easily completed; SAC can generalise to overcome features similar to its original training data. 
        
        In contrast to the SAC-trained model, the AIRL-trained model shows a median and maximum track completion of 55\% around the inverse FSG track. This corresponds to the point where two parts of the track are closest (see corner between features 2 and 3 in Figure \ref{fig:track_FSG}). Using the default 10\,m camera range here results in cones on another leg of the track being observed. These throw off the AIRL-trained model and cause it to fail, suggesting that the AIRL-trained model is more sensitive to ``random" cone observations than SAC. Reducing the camera range to 8\,m prevents this phenomenon and so AIRL sees a similar clustering of results to SAC (4 fail at 43\% and 6 complete 100\%). 

    \subsubsection{Trajectory and Smoothness}
        \begin{figure}[htbp] 
                \centering
                \includegraphics[width=0.5\textwidth]{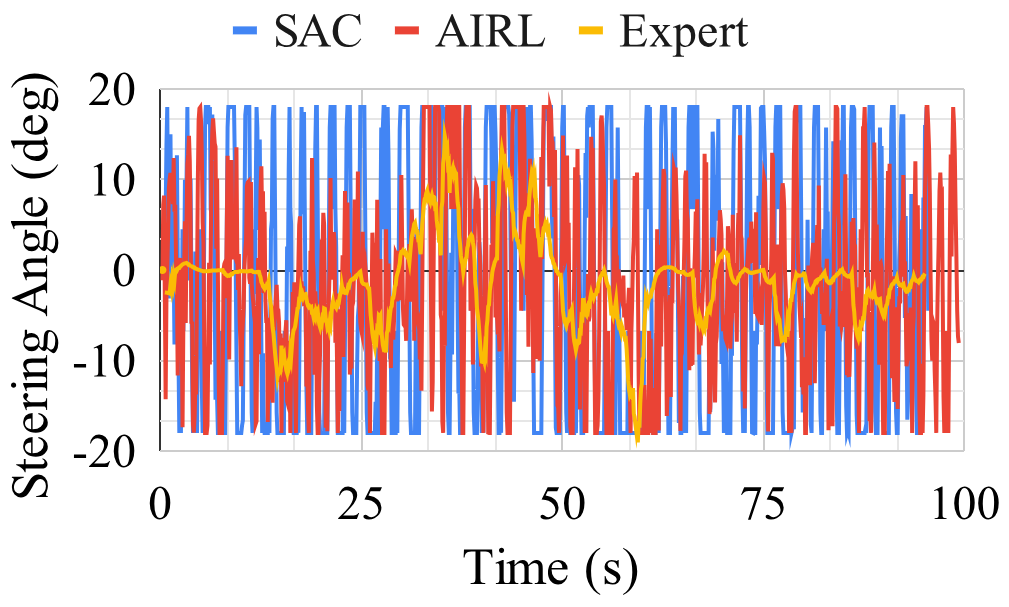}%
                \caption{SAC, AIRL and expert steering angles during an inference on the FSG track.}\label{fig:res_inf_sac_airl_FSG}
        \end{figure}
        
        Figure \ref{fig:res_inf_sac_airl_FSG} compares the steering angle selected by the expert to the RL models during a lap around the FSG track. Recall that the throttle position is selected to maintain a constant speed of $4\,m/s$, and thus the steering angle selection represents the only choice made by the RL-trained models. 
        
        It is clear that the AIRL-trained model is more similar to the expert trajectory which corresponds to less zig-zagging around the track. The mean steering angle acceleration from these graphs presented in Table \ref{table:res_steerSmooth} show that both RL models are significantly ``noisier"/less smooth than the expert: expert has mean acceleration of $4.4^{\circ}/s^2$ while SAC and AIRL have $57.3^{\circ}/s^2$ and $54.7^{\circ}/s^2$, respectively. The second column in Table \ref{table:res_steerSmooth} shows a similar trend on the FSG inverse track.

        \begin{table}[ht]
            \centering
            \begin{tabular}{l r r}\toprule
                & \multicolumn{2}{c}{Mean steering angle}\\
                & \multicolumn{2}{c}{acceleration ($deg/s^2$)}\\
                Model & FSG & Inverse FSG \\\midrule
                Expert & 4.4 & N/A \\
                SAC & 57.3 & 57.6 \\
                AIRL & 54.7 & 56.9 \\\bottomrule
                \addlinespace
            \end{tabular}
            \caption{Steering smoothness of models across different tracks.}
            \label{table:res_steerSmooth}
        \end{table}

    \subsubsection{Speed and Time-step}

        \begin{table}[htbp]
            \centering
            \includegraphics[width=0.5\textwidth]{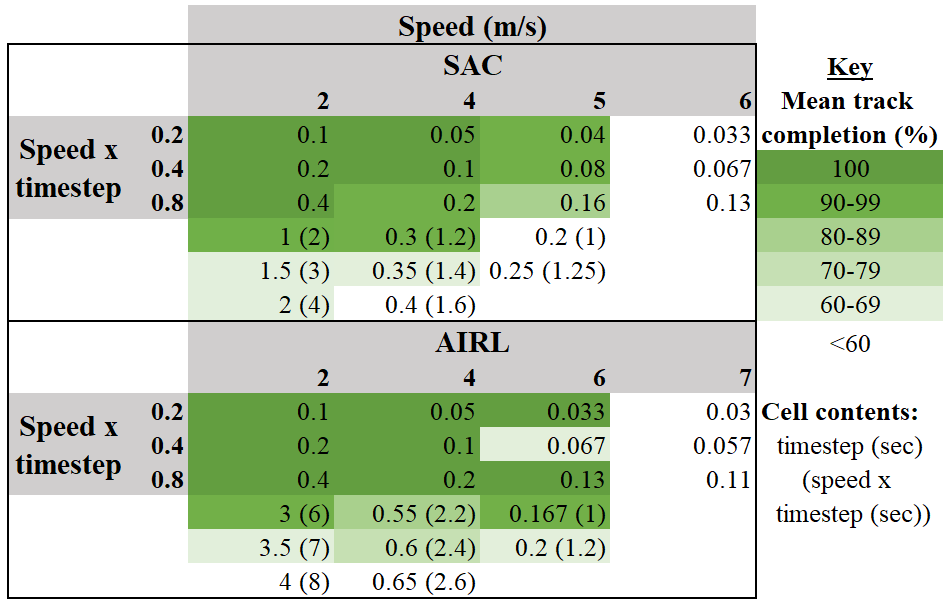}%
            \caption{Inference performance with speed and time-step variations.}\label{table:res_speed-timestep}
        \end{table}
        
        Table \ref{table:res_speed-timestep} summarizes the inference performance of the RL models across five trials, featuring various combinations of speed and time-step. This analysis aims to evaluate how the performance of the trained models fluctuates when making steering angle decisions at different rates, in relation to the vehicle's speed relative to the cones. The hypothesis was that the performance would depend on the speed\,$\times$\,time-step product; smaller decision time-steps are required to keep up with increased vehicle speeds. However, it was found that this product didn't carry much significance and rather there was simply a maximum time-step at each speed after which performance rapidly dropped off. For both SAC and AIRL, this maximum time-step reduced as speed increased. For SAC, this was from 1\,s to 0.3\,s and 0.16\,s at 2, 4 and 5\,m/s respectively and for AIRL: from 3\,s to 0.55\,s and 0.167\,s at 2, 4 and 6\,m/s respectively. Note that the AIRL model is capable of performing to a higher time-step at each speed and to a higher maximum speed (6\,m/s vs 5\,m/s) than the SAC model when both are trained at a speed of 4\,m/s and a time-step of 0.1\,s.

\section{Real World Setup and Results}
    \begin{figure}[htbp]
        \centering
        \includegraphics[width=0.25\textwidth]{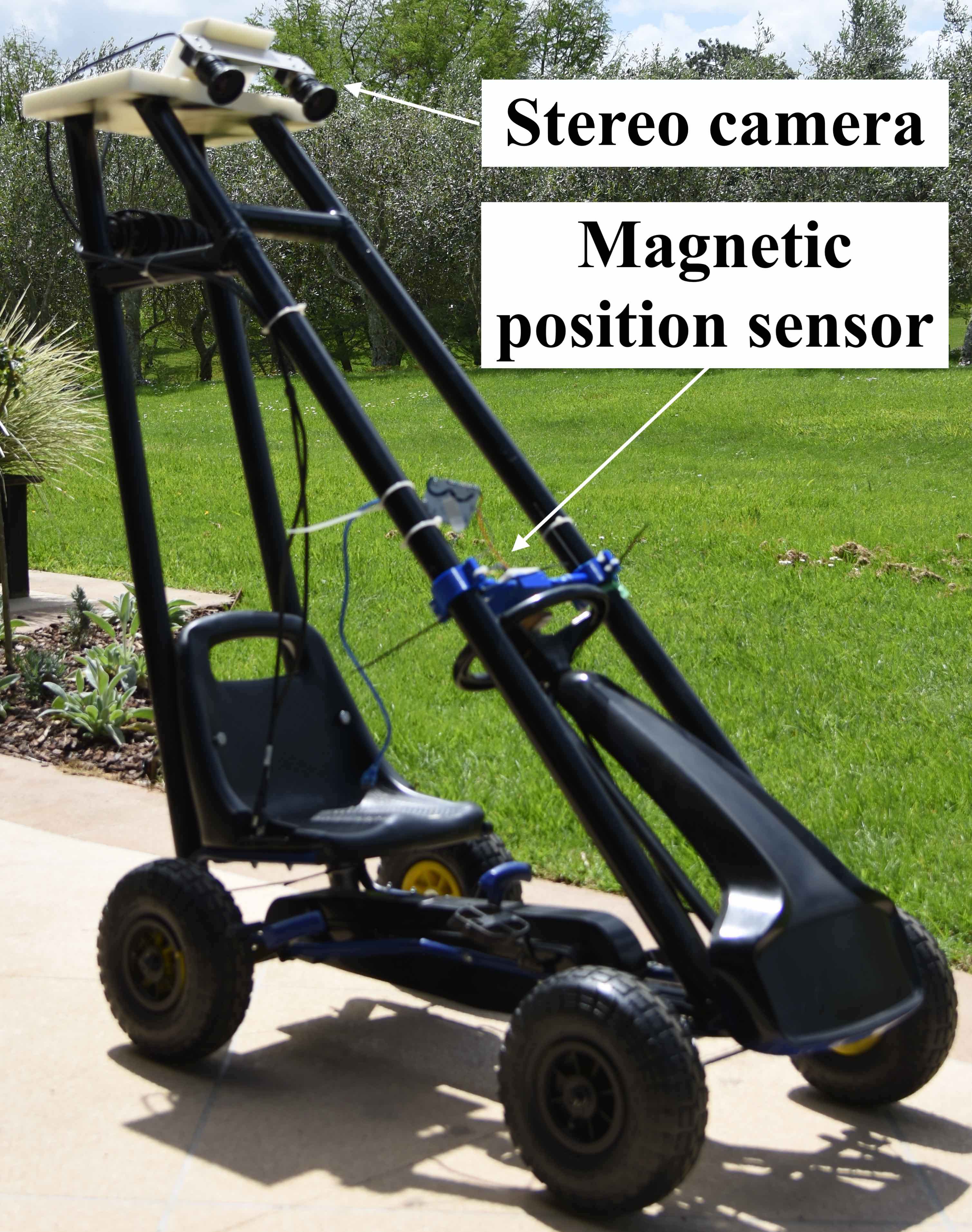}%
        \caption{Physical Testing Kart.}\label{fig:test_cart}
    \end{figure}
    
     As no functioning F:SAE vehicles were available, the system is built on a go-kart with Ackermann steering, roughly half the size of an F:SAE vehicle (see Figure \ref{fig:test_cart}). PVC pipe and a 3D printed platform is used to hold the stereo camera in a position comparable to being mounted on an F:SAE roll hoop. A magnetic rotary position sensor mounted on the steering wheel is used to observe the steering angle. 
    
\begin{figure}[htbp]
                \centering
                \includegraphics[width=0.35\textwidth]{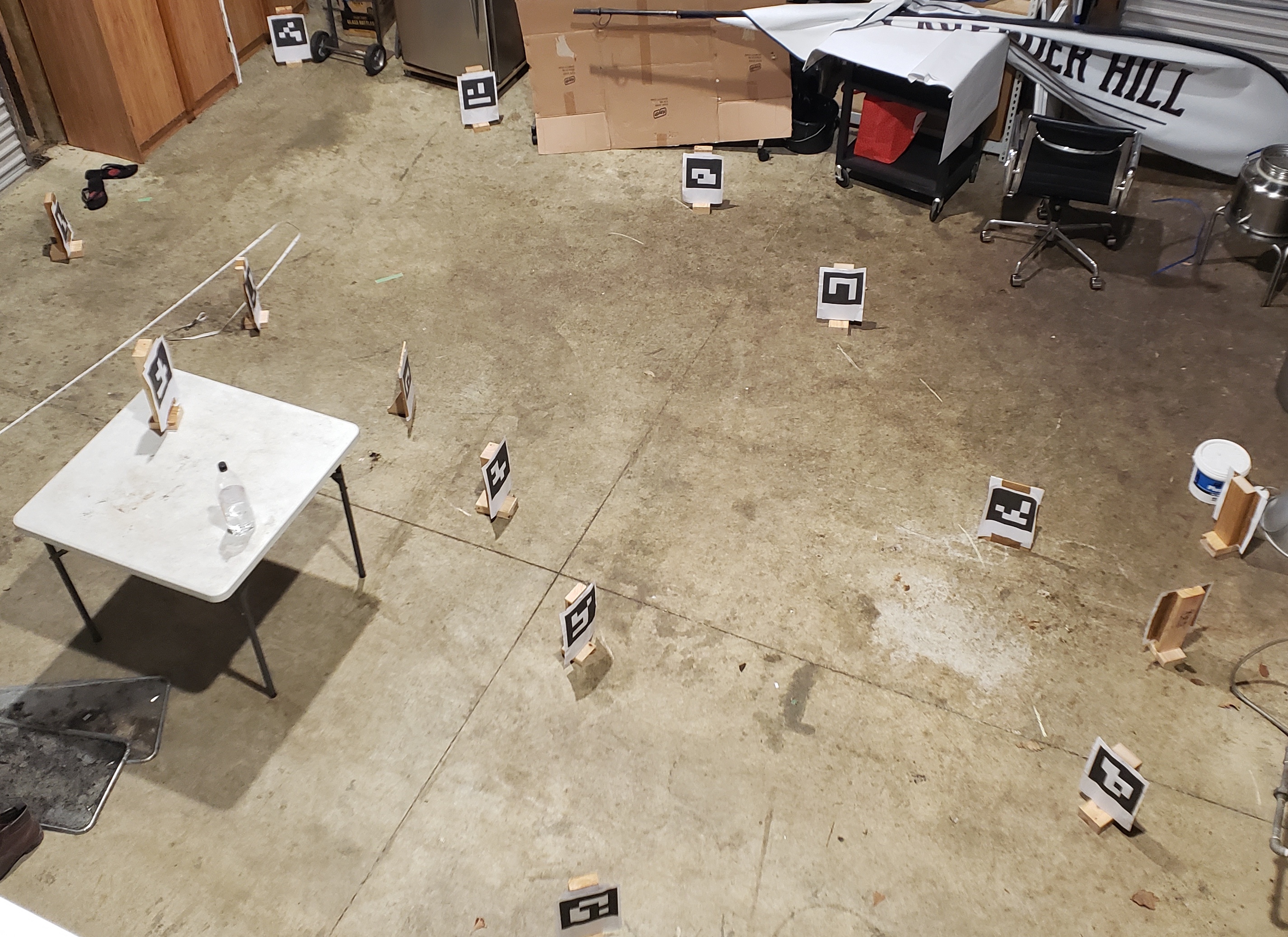}%
                \caption{Physical left turn track setup.}\label{fig:track_phys_L}
        \end{figure}

    During a test, a driver sits in the kart and is pushed around a test track (see Figure \ref{fig:track_phys_L} for an example track) at roughly 2\,m/s (estimated). As the algorithms have proven to function well between 2\,-\,5\,m/s on a full scale vehicle (1\,-\,2.5\,m/s at this half scale) with a step size of 0.1\,s, exact speed sensing is not required to keep the estimated speed within this range. 
    
    While being pushed, the driver simply turns the steering wheel in the direction and by the (degree) amount shown on a laptop's screen sitting on their lap, attempting to follow the commands as closely as possible. A blanket is placed over the front of the kart during testing to ensure the driver does not navigate using other visual cues.

    \subsubsection{Tracks}

        Tracks in the real world are marked with ArUco markers that simulate blue and yellow ``cones". The stereo camera, utilized for marker detection, is highly sensitive to variations in light intensity, necessitating frequent recalibration under dynamic outdoor lighting conditions. To address this issue, track segments were set up indoors for real world testing.
        
\begin{figure}[htbp] 
            \centering
            \includegraphics[width=0.3\textwidth]{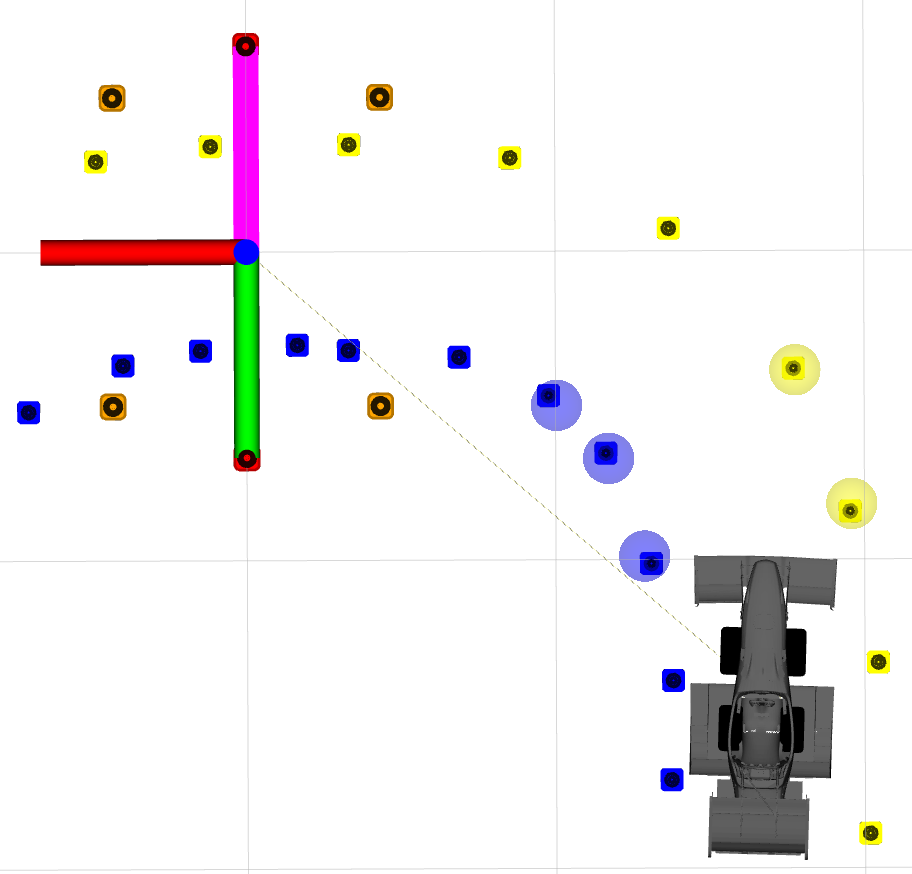}%
            \caption{Simulated left turn track setup.}\label{fig:track_sim_L}
        \end{figure}

        Four tracks are designed, each representing one of the features marked in Figure \ref{fig:track_FSG}: a straight track (6\,m), left turn ($90^{\circ}$ of a 5.5\,m radius), tight right turn ($130^{\circ}$ of a 3\,m radius) and loose right turn ($60^{\circ}$ of a 6\,m radius). As the radii of the corners and width of the tracks are approximately halved to match the kart's scale, a half scale simulation environment is also built to allow independent scaling performance analysis (see Figure \ref{fig:track_sim_L}).
        
    \subsection{Inference Results and Discussion}
\begin{figure}[htbp] 
            \centering
            \hspace*{-0.5cm}
            \includegraphics[width=0.52\textwidth]{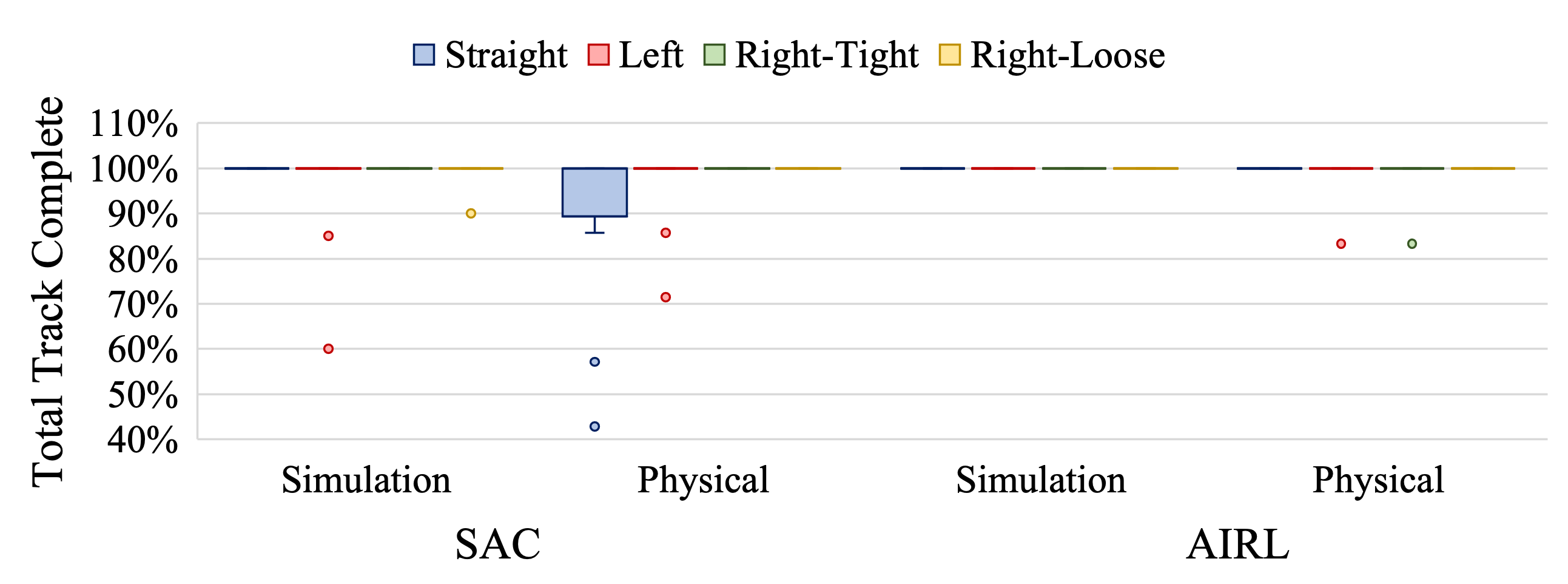}%
            \caption{SAC vs AIRL inference performance on (scaled) simulated and physical tracks.}\label{fig:res_inf_phys_repeat}
        \end{figure}

        Figure \ref{fig:res_inf_phys_repeat} compares the track completion across the four simulated (scaled) and physical tracks for both the SAC and AIRL algorithms. The results are drawn from 10 repetitions on each track. Clearly, models trained by both algorithms achieve a median completion rate of 100\% across all tracks. The AIRL-trained model proved to be slightly more repeatable when scaling down in simulation; 100\% track completion in all trials vs 92.5\% of trials achieving full completion from SAC. Similarly, the AIRL-trained model performed slightly better when transferred to the real world with 95\% of all trials achieving full completion compared to a total completion rate of 87.5\% from the SAC-trained model. Of particular note is SAC's slightly reduced reliability on the straight physical track with a lower quartile track completion rate of 90\%. The slight performance decrease in the real world is expected due to the simulation-reality gap. In this scenario, the gap is mostly due to differences in lighting, exact cone positioning and ability to follow the desired steering angle. 

        \begin{figure}[htbp] 
            \centering
            \includegraphics[width=0.4\textwidth]{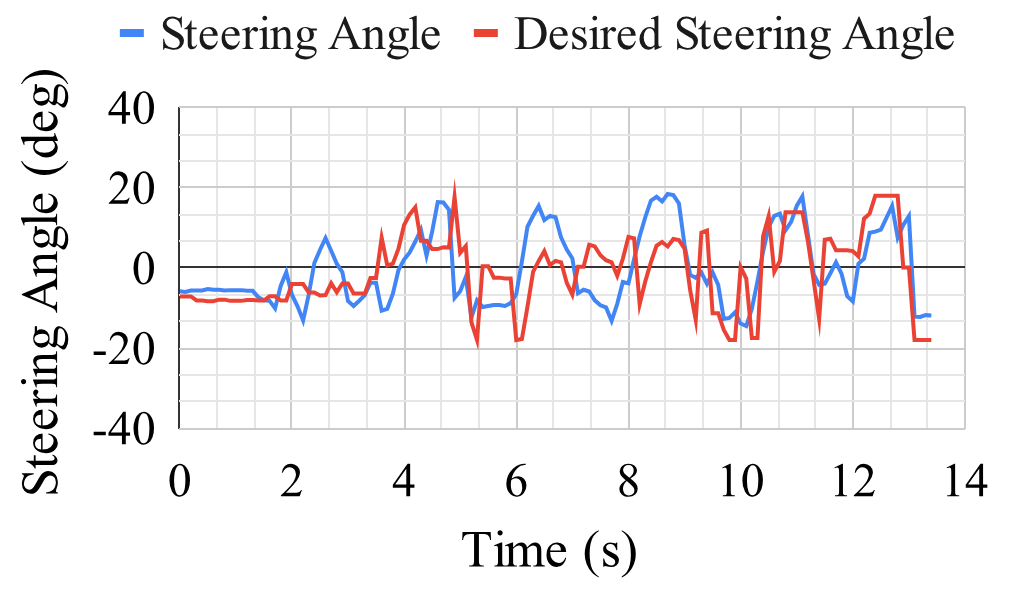}%
            \caption{SAC inference on straight physical track.}\label{fig:res_inf_phys_sac_str}
        \end{figure}
        \begin{figure}[htbp]
            \centering
            \includegraphics[width=0.4\textwidth]{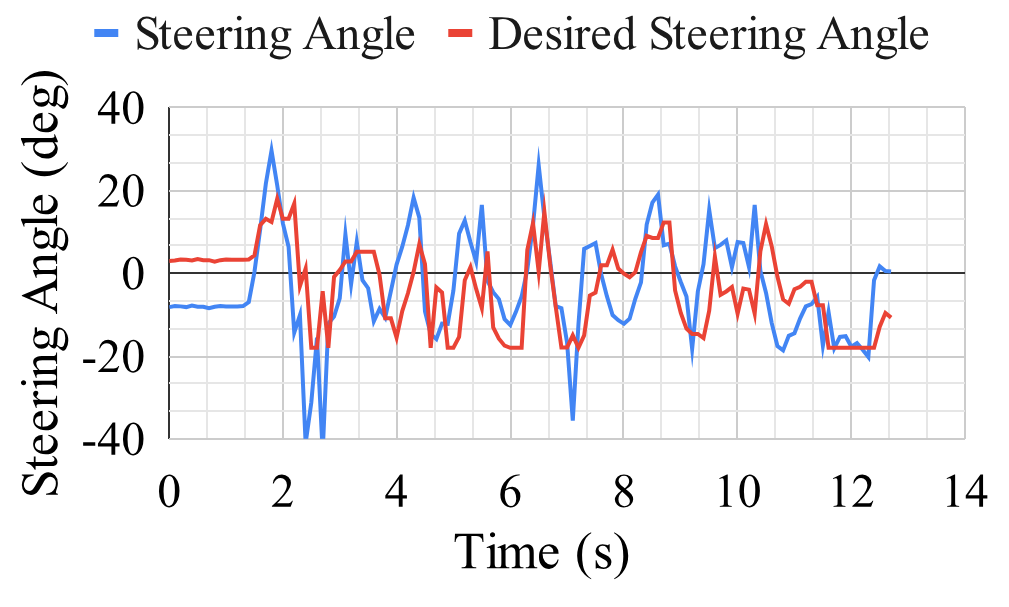}%
            \caption{AIRL inference on straight physical track.}\label{fig:res_inf_phys_airl_str}
        \end{figure}
        
        Figures \ref{fig:res_inf_phys_sac_str} and \ref{fig:res_inf_phys_airl_str} show the difference in the actual kart steering angle and the angle desired by the RL models during one trial on the straight tracks. Although the error was moderate (mean absolute error of $7.5^{\circ}$ for SAC and $9.9^{\circ}$ for AIRL), both models remained reasonably robust to this. However, the slightly lower performance of the SAC-trained model on the straight physical track may be due to lower robustness to tracking error. This error is due to the human driver's inability to perfectly follow the desired steering angle and was unavoidable as no drive-by-wire testing system, or practical alternative, was available. Note that the desired steering angle in Figures \ref{fig:res_inf_phys_sac_str} and \ref{fig:res_inf_phys_airl_str} appear smoother than those in Figure \ref{fig:res_inf_sac_airl_FSG} due to the different scales of the axes.

\section{Conclusions}
    With driverless events being introduced at F:SAE competitions, teams are beginning to investigate RL approaches. This paper focuses on the comparison of forward and inverse RL for training models capable of local path planning. The following conclusions are drawn from testing the SAC and AIRL algorithms for this purpose:

    \begin{itemize}
        \setlength\itemsep{0.25em}
        \item SAC converges faster than AIRL when training in simulation (735 vs 1386 episodes).
        \item The fastest and most physically-representative training results are achieved with steering angle limits of $\pm{18}^{\circ}$, a steering rate limit of $\pm{112.5}^{\circ}/s$ and using $\pm{1}$ as cone colour identifiers. AIRL is trained with a rollout length of 1,000 steps. 
        \item AIRL-trained models perform more reliably than SAC-trained models in simulated inference with 100\% median completion on both the training track and a different track. SAC achieves 100\% median completion on the training track and 70\% on an unknown track. 
        \item AIRL-trained models can perform at a higher speed of 6\,m/s than SAC-trained models (capped at 5\,m/s) when both are trained at the same speed (4\,m/s). Similarly, at the same speed of 4\,m/s, AIRL-trained models can function at a longer time-step of 0.55\,s vs SAC's 0.3\,s. 
        \item An AIRL-trained model performs slightly more reliably on scaled-down simulation tracks (100\% track completion rate vs 92.5\% from SAC).
        \item An AIRL-trained model performs slightly more reliably when transferred to the real world with a 95\% full track completion rate compared to 87.5\% from SAC.
    \end{itemize} 
    
    With a mean steering angle acceleration demanded from SAC of $57.6^{\circ}/s^2$ and AIRL of $56.9^{\circ}/s^2$, neither algorithm is ready for transfer to a real F:SAE vehicle (smooth expert is $4.4^{\circ}/s^2$). However, with additional work, AIRL  shows  promise  of  good  performance  in  the  shorter term,  while  SAC  has  the potential  to  outperform  both  AIRL  and  traditional  methods  in  the longer  term.

\section{Suggestions for Future Work}\label{sect:future_work}
    The main areas for further investigation are:
    
    \begin{itemize}
        \item Test training performance with varied back-end ANN parameters such as learning rate, number of hidden layers, number of neurons per hidden layer and seed weights. 
        \item Test the performance of SAC with the reward Functions \ref{eq:rewardDist} and \ref{eq:rewardTime}. 
        \item Trial common training techniques to improve robustness such as transfer learning (train with multiple tracks of different features, scales and at different speeds) and reward shaping.
        \item Allow varied vehicle speed during a lap both through relating speed to the selected steering angle and by allowing the RL algorithms to select it directly.
        \item Test algorithms on a full-scale drive-by-wire F:SAE vehicle with real competition cones marking a full-scale track. 
    \end{itemize}

\bibliographystyle{named}
\bibliography{acra}

\end{document}